\documentclass[letterpaper, 10 pt, journal, twoside]{IEEEtran}
\usepackage{amsmath,amsfonts}
\usepackage{array}
\usepackage[caption=false,font=normalsize,labelfont=sf,textfont=sf]{subfig}
\usepackage{textcomp}
\usepackage{stfloats}
\usepackage{url}
\usepackage{verbatim}
\usepackage{graphicx}
\usepackage{cite}
\usepackage{booktabs}
\usepackage{bm}
\usepackage{wrapfig}
\usepackage[ruled,vlined]{algorithm2e}
\usepackage[noend]{algpseudocode}
\usepackage{tabularx}
\usepackage{multirow, makecell}
\usepackage{amssymb}
\usepackage[table]{xcolor}
\usepackage{cancel}
\usepackage{multirow}
\usepackage{booktabs}
\usepackage{pifont}
\newcommand{\xmark}{\ding{55}}%

\newcommand{\vx}{{\bf x}}

\newcommand{\vp}{{\bf p}}

\newcommand{\vf}{{\bf f}}

\newcommand{\vu}{{\bf u}}

\newcommand{\vv}{{\bf v}}

\newcommand{\vh}{{\bf h}}

\newcommand{\vM}{{\bf M}}

\newcommand{\vomega}{{\mbox{\boldmath$\omega$}}}

\newcommand{\argmax}{\operatornamewithlimits{argmax}}

\newcommand{\E}{\mathbb{E}}

\newcommand{\ExP}[2]{\E_{{#1}}{\left[#2\right]}}

\begin{document}

\title{Learning Terrain-Aware Kinodynamic Model for Autonomous Off-Road Rally Driving With Model Predictive Path Integral Control}

\author{Hojin Lee, Taekyung Kim, Jungwi Mun, and Wonsuk Lee
\thanks{Manuscript received: April 30, 2023; Revised: August 13, 2023; Accepted: September 11, 2023. This paper was recommended for publication by Editor J. Kober upon evaluation of the Associate Editor and Reviewers' comments. This work was supported by the Agency for Defense Development Grant funded by the Korean Government in 2023. \textit{(Hojin Lee and Taekyung Kim are co-first authors.) (Corresponding author: Wonsuk Lee.)}}%
\thanks{Hojin Lee, Jungwi Mun, and Wonsuk Lee are with the AI Autonomy Technology Center, Agency for Defense Development, Daejeon, 34186, Republic of Korea {\tt\footnotesize \{hojini1117, moonjw1109, wsblues82\}@gmail.com}}%
\thanks{Taekyung Kim is with the Department of Robotics, University of Michigan, Ann Arbor, MI, 48109, USA {\tt\footnotesize taekyung@umich.edu}}%
\thanks{Digital Object Identifier (DOI): 10.1109/LRA.2023.3318190}%
}
\markboth{IEEE Robotics and Automation Letters. Preprint Version. Accepted September, 2023}%
{Lee \MakeLowercase{\textit{et al.}}: LEARNING TERRAIN-AWARE KINODYNAMIC MODEL FOR AUTONOMOUS OFF-ROAD RALLY DRIVING}


\maketitle

\begin{abstract}
High-speed autonomous driving in off-road environments has immense potential for various applications, but it also presents challenges due to the complexity of vehicle-terrain interactions. In such environments, it is crucial for the vehicle to predict its motion and adjust its controls proactively in response to environmental changes, such as variations in terrain elevation. To this end, we propose a method for learning terrain-aware kinodynamic model which is conditioned on both proprioceptive and exteroceptive information. The proposed model generates reliable predictions of 6-degree-of-freedom motion and can even estimate contact interactions without requiring ground truth force data during training. This enables the design of a safe and robust model predictive controller through appropriate cost function design which penalizes sampled trajectories with unstable motion, unsafe interactions, and high levels of uncertainty derived from the model. We demonstrate the effectiveness of our approach through experiments on a simulated off-road track, showing that our proposed model-controller pair outperforms the baseline and ensures robust high-speed driving performance without control failure. \footnote{Project page: \url{https://sites.google.com/view/terrainawarekinodyn}}
\end{abstract}


\section{Introduction}
\IEEEPARstart{A}{utonomous} navigation in unstructured environments poses a significant challenge due to the complex and unpredictable nature of such surroundings. In various applications, including surveillance and rescue missions, high-speed driving capability is crucial for successful operation, particularly in off-road environments where unpaved roads and other challenging terrains may be encountered. Therefore, the control algorithm must be designed to be aware of environmental changes and to be capable of adapting promptly to them while ensuring safety.
\begin{figure}[t]
\centering
\includegraphics[width=0.99\linewidth]{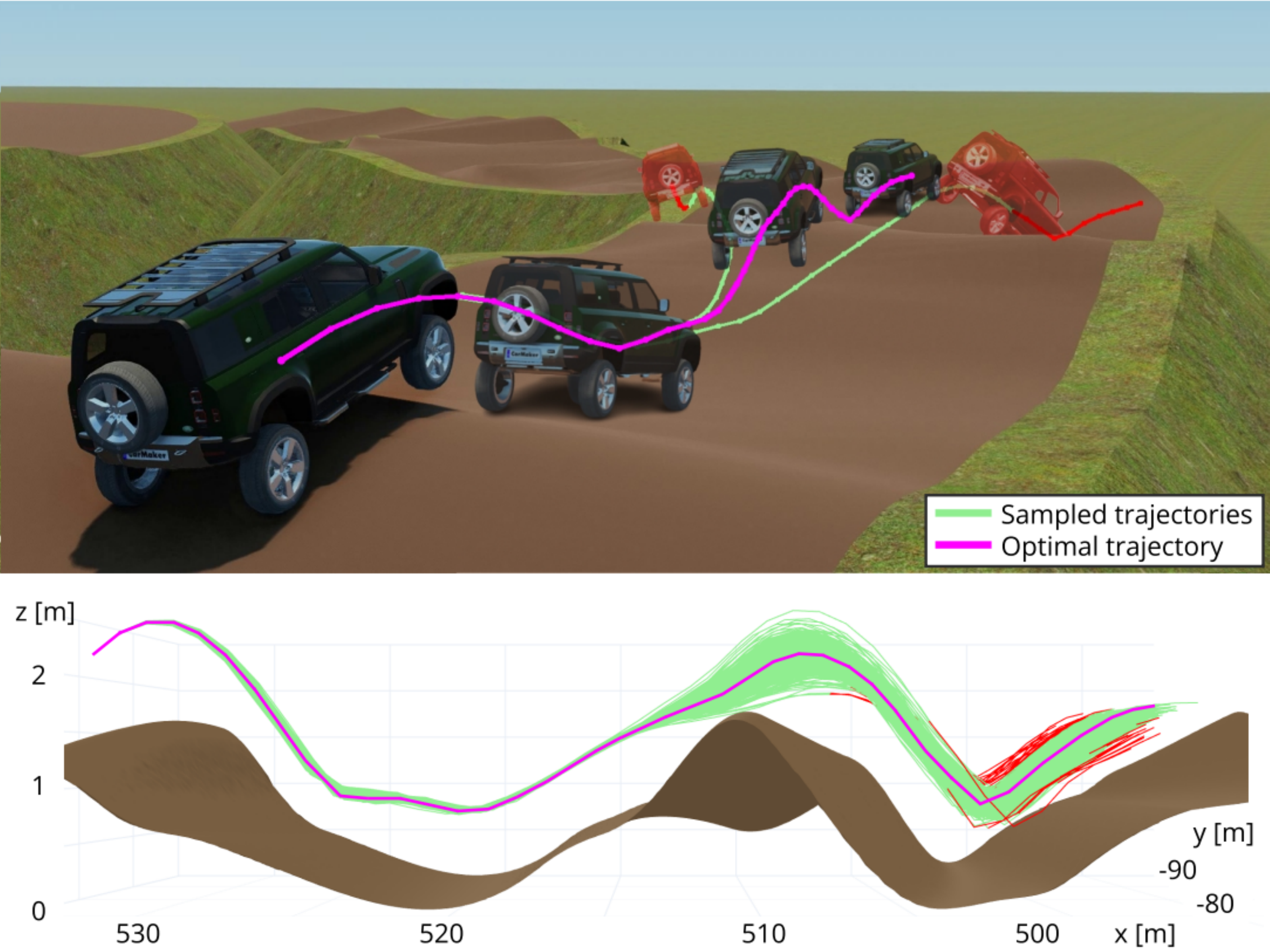}
\caption{Sampled trajectories and the optimal trajectory from the MPPI controller. Our terrain-aware kinodynamic model is capable of predicting the effect of terrain geometry on vehicle motion. The controller penalizes trajectories with unsafe vehicle motions and contact interactions, as well as high levels of uncertainty, allowing for the execution of aggressive maneuvers with enhanced safety. The penalized trajectories are depicted in red.}
\label{fig:intro_figure}
\end{figure}
\begin{figure*}[t]
\centering
\includegraphics[width=0.99\textwidth]{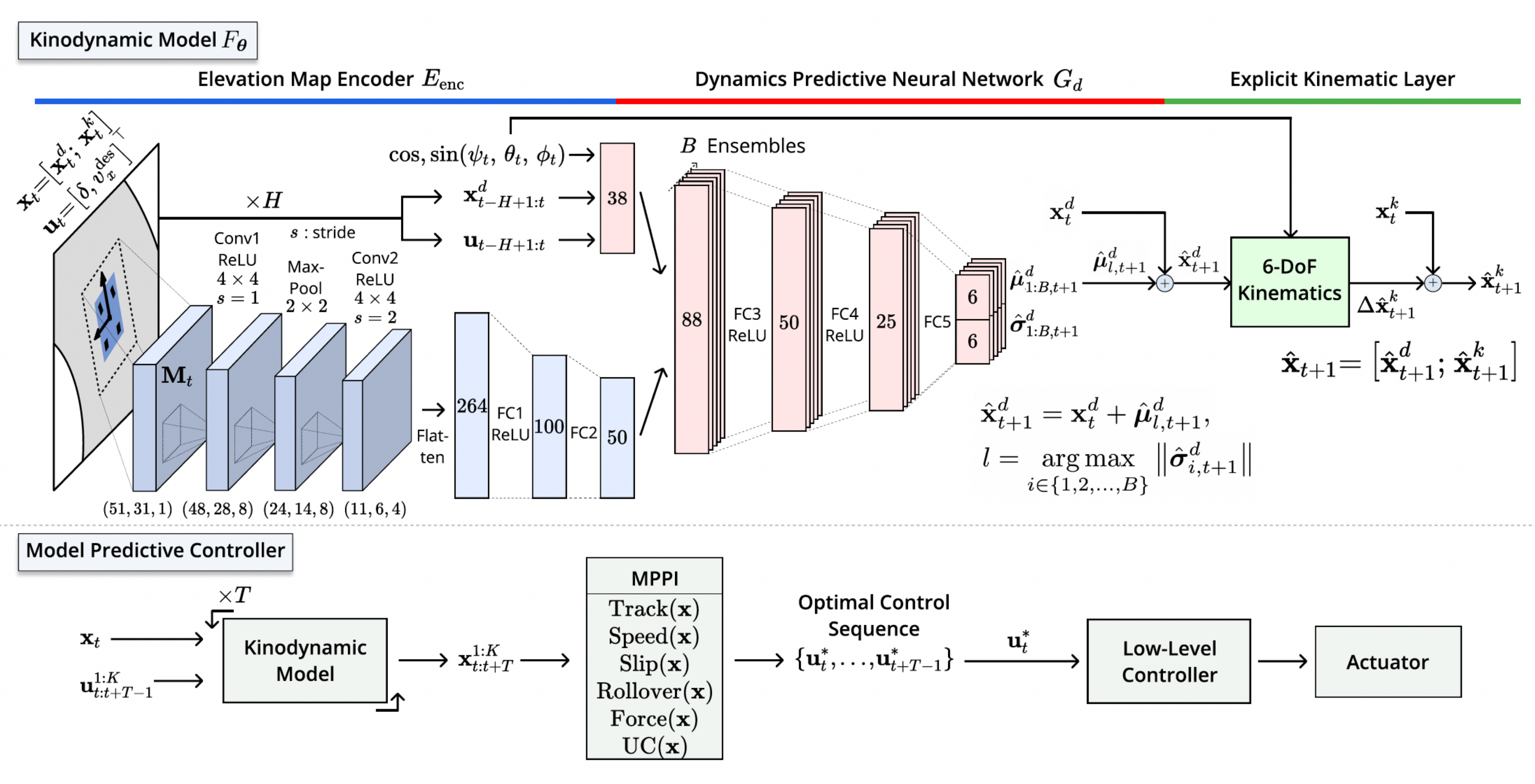}
\caption{An overview diagram of our method of learning terrain-aware kinodynamic model for autonomous high-speed off-road driving using MPPI controller.}
\label{fig:main_figure}
\end{figure*}

Although extensive research efforts have addressed high-speed autonomous driving tasks, a major portion of these studies have focused on driving under flat terrain conditions~\cite{betz2022autonomous}. Unfortunately, these approaches are not well-suited for off-road environments due to the challenges in accurately modeling the interactions between the vehicle and the terrain characterized by complex geometry. There have been attempts at analytical modeling of these interactions by using conventional methods such as with differential equations~\cite{tarokh2005kinematics, howard2007optimal}. However, these approaches rely heavily on assumptions that include constant contact between the wheels and the ground, single point contact, etc. These simplifying assumptions can be easily violated, leading to inaccurate predictions and degraded control performance, particularly in high-speed operations.

To effectively capture contact interactions in off-road environments, it is critical to model the vehicle dynamics by incorporating high-dimensional sensory data. Data-driven methods, such as Gaussian processes or online system identification, have emerged as tantalizing solutions for modeling nonlinear dynamics~\cite{khalil2002modeling, deisenroth2011pilco}. However, most of these methods rely solely on proprioceptive information and are not scalable for handling vast amounts of exteroceptive data from various sensor modalities. In recent years, researchers have shifted their focus towards neural network-based approaches that can cope with large training datasets and guarantee constant-time inference~\cite{nagabandi2018learning, sonker2020adding, kahn2021badgr, wang2021rough, tremblay2021multimodal, karnan2022vi, triest2022tartandrive}. By leveraging the advanced deep learning techniques, autonomous vehicles gain a greater understanding of their surroundings and are able to proactively adjust their control strategy, paving the way for new navigational capabilities in challenging environments.

Despite the promising outcomes of using exteroceptive information to encode environmental context in neural network dynamics, there are still a few limitations that need to be resolved. Recent studies have demonstrated the effectiveness of utilizing learning-based models as dynamics of the systems combined with model-based controllers. Nonetheless, their applications are typically limited to path following tasks, where the path is computed offline without considering wheel-terrain interactions~\cite{nagabandi2018learning, sonker2020adding, karnan2022vi}. These fail to fully harness the model's capability to reliably predict interactive motions with the environments. Concretely, it is apparent that the model-based controller has the potential to generate kinodynamically optimal future trajectories on its own by utilizing neural network dynamics as a forward dynamics model~\cite{kahn2021badgr, wang2021rough}. High-speed maneuvers in off-road environments including abrupt terrain profile changes may encompass highly discontinuous and impulsive contact forces to the systems. Pre-computed trajectories that ignore these effects can potentially degrade maneuverability, and in the worst case, lead to a loss of stability. Fig.~\ref{fig:intro_figure} illustrates an example of such a scenario, where the vehicle exhibits significant roll and pitch motions due to the vertical forces exerted on the wheels. Therefore, the controller must be capable of optimizing future trajectories that are inherently feasible and robust to the environmental changes, respecting both the structural shape of the terrain and the dynamic characteristics of the controlled system.

In this paper, we propose an approach of learning terrain-aware kinodynamic model to achieve high-speed vehicle control in off-road environments while ensuring safety and robustness. Our contributions are as follows:
\begin{itemize}
    \item We develop a neural network model that can make reliable 6 Degrees-of-Freedom~(DoF) motion predictions in terrains with rapid profile changes, by using a local elevation map as input and adopting a number of design considerations. In addition, our model can estimate the severity of ground impact by deriving contact forces and moments from its motion predictions, even without requiring ground truth force data during training.
    \item We describe a sampling-based model predictive control~(MPC) scheme that employs our model as the forward dynamics to directly plan kinodynamically optimal trajectories on its own. This is achieved by a cost function design which penalizes trajectories that exhibit unstable motion, unsafe interactions, and high levels of uncertainty, all of which can be inferred from our model. 
    \item We demonstrate the effectiveness of our idea by conducting experiments on a simulated off-road race track. Additionally, we validate that our method enhances vehicle maneuverability to a greater extent than the baseline, by presenting a scenario where our model encourages the MPC to make aggressive maneuvers rather than taking a conservative behavior.
\end{itemize}

\section{Related Work}
Early work on off-road autonomous driving mainly centered on analyzing the effect of vehicle-terrain interaction via traversability estimation~\cite{krusi2017driving, chavez2018learning, maturana2018real}. They primarily used human supervision to assess the difficulty of the terrain to be traversed and to provide the vehicle with online maps of traversable regions. More recently, researchers have focused on self-supervised approaches that aim to learn traversability directly from real driving experience~\cite{sathyamoorthy2022terrapn, seo2023scate}. Unfortunately, all these methods may suffer from dynamics-related information loss when dealing with challenging terrain, because they tend to simplify traversability as a scalar value or a categorical label, and do not consider dynamic interactions which vary depending on the dynamic state of the vehicle.

On the other hand, a dynamics learning approach for off-road driving has been developed with the aim of capturing both the nonlinear vehicle dynamics and its interaction with the terrain, without relying on simplified traversability representations. Xiao et al.~\cite{xiao2021learning} developed an inverse kinodynamic model conditioned on onboard proprioceptive observations to sense the effect of vehicle-terrain interaction. Other previous studies demonstrated the benefits of leveraging exteroceptive information along with proprioceptive information to prepare the vehicle for upcoming interactions~\cite{nagabandi2018learning, sonker2020adding, karnan2022vi}. Building upon these strategies, Tremblay et al.~\cite{tremblay2021multimodal} and Triest et al.~\cite{triest2022tartandrive} have further improved off-road dynamics prediction by expanding the model to accommodate multiple high-dimensional sensor modalities.

An off-road dynamics model can then be used to provide useful information to a model-based controller, which can facilitate the planning of optimal trajectories. Kahn et al.~\cite{kahn2021badgr} demonstrated that the predictive model can be used to execute control inputs that prevent collisions and actively navigate the robot towards smoother terrain. Wang et al.~\cite{wang2021rough} used constrained optimization to find trajectories with low divergence inferred from a probabilistic dynamics model. In this work, we extend this control strategy to high-speed off-road driving, by enabling the controller to perceive unstable motion and unsafe interactions, and by optimizing control inputs to prevent catastrophic failures when driving through challenging terrain.

\section{Terrain-Aware Kinodynamic Model}
In this section, we first define the state and action space to establish the kinodynamic model~$F_{\bm{\theta}}$. It predicts the change in the vehicle's state induced by control inputs and contact interactions during a discrete time period of~$\Delta t$. Then, we describe the three main components of the model: 1) \textit{Elevation Map Encoder}, 2) \textit{Dynamics Predictive Neural Network}, and 3) \textit{Explicit Kinematic Layer}.

 First, we define the 6-DoF proprioceptive state~$\mathbf{x} = \left[\mathbf{x}^{d};\,\mathbf{x}^{k}\right]$ at the center of mass~(CoM) of the vehicle. The dynamic state~$\mathbf{x}^{d}$ includes linear velocity~$\vv_{\mathrm{b}}$ and angular velocity~$\vomega_{\mathrm{b}}$. The kinematic state~$\mathbf{x}^{k}$ comprises global position~$\vp_{\mathrm{s}}$ and orientation~$\mathbf{e}_{\mathrm{s}}$. The notations $(\cdot)_\mathrm{s}$ and $(\cdot)_\mathrm{b}$ denote vectors represented in a fixed reference frame and in the vehicle body frame attached to the CoM, respectively:
\begin{equation}
\begin{aligned}
     \mathbf{x}^{d} & =
     \begin{bmatrix}
    \vv_{\mathrm{b}}\\
    \vomega_{\mathrm{b}}
    \end{bmatrix} = \left[v_{x},\, v_{y},\, v_{z},\, \omega_{x},\, \omega_{y},\, \omega_{z}\right]^{\top}, \\
     \mathbf{x}^{k} & =
    \begin{bmatrix}
    \vp_{\mathrm{s}}\\
    \mathbf{e}_{\mathrm{s}}
    \end{bmatrix}
    = \left[x,\, y,\, z,\, \psi,\, \theta,\, \phi\right]^{\top}, 
\label{Equation:proprioceptive_state}
\end{aligned}
\end{equation}
where $\left(\psi, \theta, \phi\right)$ are the yaw, pitch, and roll of the ZYX Euler angles, respectively.

Second, we augment the state space with exteroceptive information to incorporate the influence of terrain geometry. In a recent study, exteroceptive information consisted of height samples obtained near each foot of the robot~\cite{miki2022learning}. For bipedal or quadrupedal robots, contact points can be computed precisely using rigid body kinematics~\cite{wellhausen2019should}. However, for full-size vehicles, deformable suspensions make the estimation of contact interaction unreliable using height samples around computed contact points.

Therefore, we use a vehicle-centric local elevation map cropped from the global map as exteroceptive information. We assume that the vehicle has full knowledge of the global map through existing techniques~\cite{fankhauser2018probabilistic}. To ensure the model captures all possible contact interactions within~$\Delta t$, we first specify the maximum operating speed of the vehicle, and set the size of the local map to be large enough to encompass the area that the vehicle can reach within~$\Delta t$ from its current position, as well as all four contact points. This strategy enables the neural network to implicitly predict contacts and interactions by considering the influence of terrain geometry around the vehicle as a whole, as opposed to using only the local geometry at each contact point. The local map is centered at the CoM and aligned with the yaw angle of the vehicle. We represent the elevation magnitudes of the local map, denoted as~$\vM$, as relative heights compared to the CoM. We set~$\Delta t$ to $0.1$~s and the grid size of the local map to $10$~cm.

In contrast to the electric vehicles used in recent learning-based autonomous driving studies~\cite{williams2018information, kabzan2019learning}, our vehicle uses an automatic transmission for the gear-shifting mechanism, making throttle control impractical. Therefore, we select the desired speed as a control input instead of the throttle and use a Proportional-Integral~(PI) low-level controller to track the desired speed. Hence, the control input~$\vu$ consists of the steering angle~$\delta$ and the desired longitudinal speed~$v^{\text{des}}_{x}$.
\subsection{Elevation Map Encoder}
The \textit{Elevation Map Encoder}~(EME)~$E_{\text{enc}}$ processes the local elevation map~$\vM_{t}$ and outputs the latent terrain feature vector~$\vh_{t}$. $\vM_{t}$ is fed through a three-layer Convolutional Neural Network~(CNN), whose output is flattened and then fed through two additional fully-connected layers:
\begin{equation}
    \vh_{t} = E_{\text{enc}}\left(\vM_{t}\right).
\end{equation}
\subsection{Dynamics Predictive Neural Network\label{section:dpnn}}
The \textit{Dynamics Predictive Neural Network}~(DPNN)~$G_{d}$ predicts the change in linear and angular velocities using proprioceptive information and a terrain feature vector~$\mathbf{h}_{t}$ as inputs. To enhance the model's prediction performance, we take a number of design considerations for the DPNN.

In real scenarios, exteroceptive sensors often provide noisy and imperfect data about the terrain. To account for this uncertainty, we build the DPNN as a probabilistic ensemble~(PE) model~\cite{chua2018deep, kim2023bridging} capable of capturing both aleatoric and epistemic uncertainty, which can then be managed by the controller. Each ensemble member with the same network structure is randomly initialized and individually trained to capture epistemic uncertainty via disagreement in prediction. Each member outputs the mean and standard deviation vector~$\left(\hat{\boldsymbol{\mu}}^{d}_{t+1}, \,\hat{\boldsymbol{\sigma}}^{d}_{t+1}\right)$ of the normal distribution to predict the change in velocities while incorporating aleatoric uncertainty.

In addition, rapid changes in the terrain profile cause highly time-varying and nonlinear vehicle motion that is difficult to model. Prior work suggested that adding history information to neural network models is effective for capturing time-varying behavior of the vehicle~\cite{spielberg2019neural}. Therefore, we provide the DPNN with the history of velocities and control inputs~$\left(\mathbf{x}^{d}_{t-H+1:t},\, \mathbf{u}_{t-H+1:t}\right) = \left(\mathbf{x}^{d}_{t-H+1},\dots,\mathbf{x}^{d}_{t-1},\, \mathbf{x}^{d}_{t},\, \mathbf{u}_{t-H+1},\dots,\mathbf{u}_{t-1},\, \mathbf{u}_{t}\right)$. In order to capture contact interactions, we also provide orientation angles~$(\psi_t, \theta_t, \phi_t)$ to characterize the spatial relationship between the vehicle and terrain. $\cos$ and $\sin$ of angular values are used to avoid the angle wrapping problem~\cite{nagabandi2018learning}.
 
 To sum up, the DPNN can be described as:
\begin{equation}
\begin{aligned}
    \hat{\boldsymbol{\mu}}^{d}_{i, t+1},\, \hat{\boldsymbol{\sigma}}^{d}_{i, t+1} &=
    G_{d}\left(\mathbf{x}^{d}_{t-H+1:t},\, \mathbf{u}_{t-H+1:t},\, h_{t},\right. \\
    &\left. \mathrel{\phantom{(}} \hspace{0.8cm}  \mathrm{c}_{\psi} ,\, \mathrm{s}_{\psi},\, \mathrm{c}_{\theta},\, \mathrm{s}_{\theta},\, \mathrm{c}_{\phi},\, \mathrm{s}_{\phi}\right), \\
    \hat{\mathbf{x}}^{d}_{i, t+1} &= \mathbf{x}^{d}_{t} + 
    \left(\hat{\boldsymbol{\mu}}^{d}_{i, t+1} + 
    \hat{\boldsymbol{\sigma}}^{d}_{i, t+1}
    \boldsymbol{\varepsilon}_i\right), \\
    \boldsymbol{\varepsilon}_i &\sim \mathcal{N}(\mathbf{0}, \mathbf{I})\,\forall i \in \{1, 2, \dots, B\}.
\label{Equation:dpnn}
\end{aligned}
\end{equation}
For notational simplicity, we abbreviate $\cos(*)$ and $\sin(*)$ as $\mathrm{c}_{*}$ and $\mathrm{s}_{*}$, respectively. $B$ is the number of models in the ensemble. In Section~\ref{section:imp_detail}, we provide a slightly modified version of~(\ref{Equation:dpnn}) for use in the controller.
\subsection{Explicit Kinematic Layer}
The \textit{Explicit Kinematic Layer}~(EKL) analytically calculates the changes in global position and orientation. It transforms velocity outputs~$\left(\hat{\vv}_{\mathrm{b}},\,\hat{\vomega}_{\mathrm{b}}\right) \in \hat{\mathbf{x}}^{d}$ of the DPNN to vectors in a fixed reference frame. The change in position~$\Delta \hat{\vp}_{\mathrm{s}}$ can be obtained by sequentially applying the rotation matrices~$R$ corresponding to each Euler angle to the body frame displacement~$\hat{\vv}_{\mathrm{b}}\Delta t$:
\begin{equation}
\begin{aligned}
    \Delta \hat{\vp}_{\mathrm{s}} 
    & = 
    R_{z}(\psi)R_{y}(\theta)R_{x}(\phi)
    \hat{\vv}_{\mathrm{b}}\Delta t\\
    & =
    \begin{bmatrix}
    \mathrm{c}_{\psi}\mathrm{c}_{\theta} & 
    \mathrm{c}_{\psi}\mathrm{s}_{\theta}\mathrm{s}_{\phi}- \mathrm{s}_{\psi}\mathrm{c}_{\phi} & 
    \mathrm{c}_{\psi}\mathrm{s}_{\theta}\mathrm{c}_{\phi}+ \mathrm{s}_{\psi}\mathrm{s}_{\phi} \\
    \mathrm{s}_{\psi}\mathrm{c}_{\theta} & 
    \mathrm{s}_{\psi}\mathrm{s}_{\theta}\mathrm{s}_{\phi}+ \mathrm{c}_{\psi}\mathrm{c}_{\phi} &
    \mathrm{s}_{\psi}\mathrm{s}_{\theta}\mathrm{c}_{\phi}- \mathrm{c}_{\psi}\mathrm{s}_{\phi} \\
    -\mathrm{s}_{\theta} &
    \mathrm{c}_{\theta}\mathrm{s}_{\phi} & 
    \mathrm{c}_{\theta}\mathrm{c}_{\phi}
    \end{bmatrix}
    \begin{bmatrix}
        v_x\\ v_y\\ v_z
    \end{bmatrix}\Delta t.
\label{Equation:linvel_to_pos}
\end{aligned}
\end{equation}
The angular velocity vector~$\hat{\vomega}_{\mathrm{b}}$ is the sum of Euler angle rate vectors transformed from their intermediate frames to the vehicle body frame~\cite{howard2007optimal}: 
\begin{equation}
\begin{aligned}
    \hat{\vomega}_{\mathrm{b}}
    & = 
    \begin{bmatrix}
        \Dot{\phi}\\ 0\\ 0
    \end{bmatrix} +
    R_{x}(-\phi)
    \begin{bmatrix}
        0\\ \Dot{\theta}\\ 0
    \end{bmatrix} + 
    R_{x}(-\phi)R_{y}(-\theta)
    \begin{bmatrix}
        0\\ 0\\ \Dot{\psi}
    \end{bmatrix} \\
    & =
    \begin{bmatrix}
    -\mathrm{s}_{\theta} & 0 & 1 \\
    \mathrm{c}_{\theta}\mathrm{s}_{\phi} & \mathrm{c}_{\phi} &
    0 \\
    \mathrm{c}_{\theta}\mathrm{c}_{\phi} & -\mathrm{s}_{\phi} & 
    0
    \end{bmatrix}
    \begin{bmatrix}
        \Dot{\psi}\\ \Dot{\theta}\\
        \Dot{\phi}
    \end{bmatrix}.
\label{Equation:ori_to_angvel}
\end{aligned}
\end{equation}
Then the change in orientation~$\Delta \hat{\mathbf{e}}_{\mathrm{s}}$ can be derived via the inverse mapping of~(\ref{Equation:ori_to_angvel}):
\begin{equation}
       \Delta \hat{\mathbf{e}}_{\mathrm{s}}
    =     \begin{bmatrix}
        \Dot{\psi}\\ \Dot{\theta}\\
        \Dot{\phi}
    \end{bmatrix} \Delta t = 
    \begin{bmatrix}
    0 & \mathrm{s}_{\phi}/\mathrm{c}_{\theta} &
    \mathrm{c}_{\phi}/\mathrm{c}_{\theta} \\
    0 & \mathrm{c}_{\phi} & -\mathrm{s}_{\phi} \\
    1 & \mathrm{s}_{\theta}\mathrm{s}_{\phi}/\mathrm{c}_{\theta} & 
    \mathrm{s}_{\theta}\mathrm{c}_{\phi}/\mathrm{c}_{\theta}
    \end{bmatrix}\hat{\vomega}_{\mathrm{b}}\Delta t.
\label{Equation:angvel_to_ori}
\end{equation}
Finally, the 6-DoF kinematic state~$\hat{\mathbf{x}}_{t+1}^{k}$ is computed as:
\begin{equation}
    \hat{\mathbf{x}}^{k}_{t+1} = \mathbf{x}^{k}_{t} + 
    \begin{bmatrix}
            \Delta \hat{\vp}_{\mathrm{s}}\\     \Delta \hat{\mathbf{e}}_{\mathrm{s}}
    \end{bmatrix}_{t+1}.
\end{equation}

\section{Model Predictive Control Using The Terrain-Aware Model}
\subsection{Model Predictive Path Integral Controller}
 Here, we provide a brief overview of our MPC method for off-road driving. Sampling-based MPC can handle both convex and non-convex optimization problems without requiring a gradient of the cost function. In recent years, the MPPI algorithm has been one of the most effective sampling-based approaches, aided by advancements in Graphics Processing Units~(GPUs). The MPPI enables the design of arbitrary cost functions and the representation of the dynamics model by any nonlinear functions, such as neural networks~\cite{williams2018information}.
 
 In each timestep~$t$, $K$~parallel samples of control sequences~$\mathbf{u}^{1:K}_{t:t+T-1} = \left(\mathbf{u}^{1}_{t:t+T-1},\dots,\mathbf{u}^{K}_{t:t+T-1}\right)$ are drawn with a finite time horizon~$T$. Each control sequence sample is transformed into the corresponding state trajectory by recursively applying~$F_{\boldsymbol{\theta}}$. Then the optimal sequence~$\mathbf{u}^{*}_{t:t+T-1}$ is obtained by taking the weighted average of samples, where the weight is determined by the evaluated cost of each trajectory using state-dependent cost function~$c(\vx)$.

 However, significant chattering of the optimal sequence may occur due to the stochastic process of this sampling-based algorithm. To resolve this problem, we utilize the Smooth MPPI~(SMPPI) approach proposed in our prior work~\cite{kim2022smooth}, which effectively attenuates chattering while adapting to a rapidly changing environment. Similar to prior work~\cite{williams2018information}, we add the following three components to~$c(\vx)$: 
\begin{equation}
\begin{aligned}
    \text{Track}(\vx) & = {W_{\text{inf}} \, \mathbf{C}(x,y)} , \\
    \text{Speed}(\vx) & = (v_x - v_{\text{ref}})^2 , \\
    \text{Slip}(\vx) & = \beta^2 + W_{\text{inf}}I \, ({\left | \beta \right | > \beta_{th}}).
    \label{Equation:2D_cost}
\end{aligned}
\end{equation}
``$th$" denotes an empirically defined threshold, and $\beta = -\text{tan}^{-1}\left ( \frac{v_y}{\lvert v_x \rvert} \right )$ is the side-slip angle. $\mathbf{C}(x,y)$ is the value in 2D cost-map at~$(x, y)$ which returns 1 if the vehicle is outside the track boundary and 0 otherwise. $v_{\text{ref}}$ is the predefined reference speed, and $I$ is an indicator function. $W_{\text{inf}}$ is a sufficiently large constant to impose an impulse penalty.
\subsection{Off-Road Cost Function Design}
We suggest a number of off-road cost function terms that can be incorporated into the state cost~$c(\mathbf{x})$ by employing predictions of the model. Instead of relying solely on~(\ref{Equation:2D_cost}), the inclusion of off-road costs allows the controller to make aggressive terrain-aware behavior while ensuring safety.

Firstly, we define the rollover cost to simply penalizing trajectories with large roll and pitch angles:
\begin{equation}
\begin{aligned}
    \text{Rollover}(\vx) &=  \theta^2 + \phi^2 + \\ &W_{\text{inf}}(I \, (\left | \theta \right | > \theta_{th})+ I \, (\left | \phi \right | > \phi_{th}))).
\end{aligned}
\end{equation}

Secondly, we penalize trajectories in which the vehicle is subjected to substantial interaction forces and moments resulting from terrain geometry. When driving at high speeds on off-road terrain, unstable conditions of the vehicle not only include a high roll or pitch angle, but also a large vertical force~$F_{z}$ caused by shock impacts, a roll moment~$M_{x}$ and a pitch moment~$M_{y}$ resulting from uneven force distribution. This may severely damage the vehicle or compromise its stability. We can derive interaction forces~$\hat{\vf}_{\mathrm{b}}$ and moments~$\hat{\mathbf{m}}_{\mathrm{b}}$ using velocity outputs from our model as:
\begin{equation}
\begin{aligned}
    \hat{\vf}_{\mathrm{b}} &= \begin{bmatrix}
    F_{x}\\ F_{y}\\ F_{z}
\end{bmatrix}_{\mathrm{b}} = m\left(\Dot{\hat{\mathbf{v}}}_{\mathrm{b}} + 
    \begin{bmatrix}
    \hat{\vomega}_{\mathrm{b}}
    \end{bmatrix}\hat{\mathbf{v}}_{\mathrm{b}}\right), \\
    \hat{\mathbf{m}}_{\mathrm{b}} &= \begin{bmatrix}
    M_{x}\\ M_{y}\\ M_{z}
\end{bmatrix}_{\mathrm{b}} = \mathcal{I}_{\mathrm{b}}\Dot{\hat{\vomega}}_{\mathrm{b}} +     \begin{bmatrix}
    \hat{\vomega}_{\mathrm{b}}
    \end{bmatrix}\mathcal{I}_{\mathrm{b}}\hat{\vomega}_{\mathrm{b}}.
\label{Equation:force_and_moment}
\end{aligned}
\end{equation}
$\begin{bmatrix}
    \hat{\vomega}_{\mathrm{b}}
\end{bmatrix}$ is a $3 \times 3$ skew-symmetric matrix expression of~$\hat{\vomega}_{\mathrm{b}}$~\cite{lynch2017modern}. $\mathcal{I}_{\mathrm{b}}$ is the rotational inertia matrix of the vehicle. In our model, we can approximate~$\Dot{\hat{\mathbf{v}}}_{\mathrm{b}}$ as~$\left\{\hat{\mathbf{v}}_{\mathrm{b}}(t+1)-\hat{\mathbf{v}}_{\mathrm{b}}(t)\right\} / \Delta t$, and a similar approximation can be made for~$\Dot{\hat{\vomega}}_{\mathrm{b}}$. Hence we can define the force cost as follows to explicitly penalize unsafe interactions without resorting to heuristics:
\begin{equation}
\begin{aligned}
    &\text{Force}(\vx) = \sum_{j=1}^{3} \left(\zeta_{j}^{2} + W_{\text{inf}}I \, (\lvert \zeta_{j} \rvert > \zeta_{j, th})\right), \\  
    &\zeta_{1} = \frac{F_z}{mg} - 1,\, \zeta_{2} = \frac{M_x}{mgd_{x}},\, \zeta_{3} = \frac{M_y}{mgd_{y}}.
\end{aligned}
\end{equation}
$(d_{x}, d_{y})$ denotes the distances from the CoM to the front-left wheel in the lateral and longitudinal directions, respectively.
 
Lastly, we incorporate the uncertainty cost into~$c(\vx)$. This enables the controller to avoid state spaces with high uncertainty owing to noisy exteroceptive input or insufficient model learning. $\text{UC}(\vx)$ is defined as the total amount of uncertainty inherent in the model prediction, which can be expressed as the sum of aleatoric uncertainty~$\text{AU}(\vx)$ and epistemic uncertainty~$\text{EU}(\vx)$ as follows~\cite{egele2022autodeuq}:
\begin{equation}
    \text{UC}(\vx) = \text{AU}(\vx) + \text{EU}(\vx)
    = \ExP{i}{\left\lVert\hat{\boldsymbol{\sigma}}_{i}^{d}\right\rVert^{2}} + \left\lVert\operatorname{Var}_{i}\left[\hat{\boldsymbol{\mu}}_{i}^{d}\right]\right\rVert.
\end{equation}

To sum up, the final form of~$c(\vx)$ is:
\begin{equation}
\begin{aligned}
    c(\vx) &= w_{1}\text{Track}(\vx) + w_{2}\text{Speed}(\vx) + w_{3}\text{Slip}(\vx) \\ &+ w_{4}\text{Rollover}(\vx) + w_{5}\text{Force}(\vx) + w_{6}\text{UC}(\vx).
\end{aligned}
\end{equation}
\subsection{Implementation Details \label{section:imp_detail}}
This subsection explains a few details for handling predictions of our model to reduce computation and enhance robustness of the MPPI. Since the PE model provides a distribution of trajectories instead of a single trajectory, multiple trajectories based on the Monte Carlo~(MC) method are necessary to evaluate the cost of each control sequence. Such an approach requires an exponentially increasing number of state cost evaluations along sampling over each distribution, which is intractable to be implemented in real time.

\begin{figure}[t]
\centering
\includegraphics[width=0.99\linewidth]{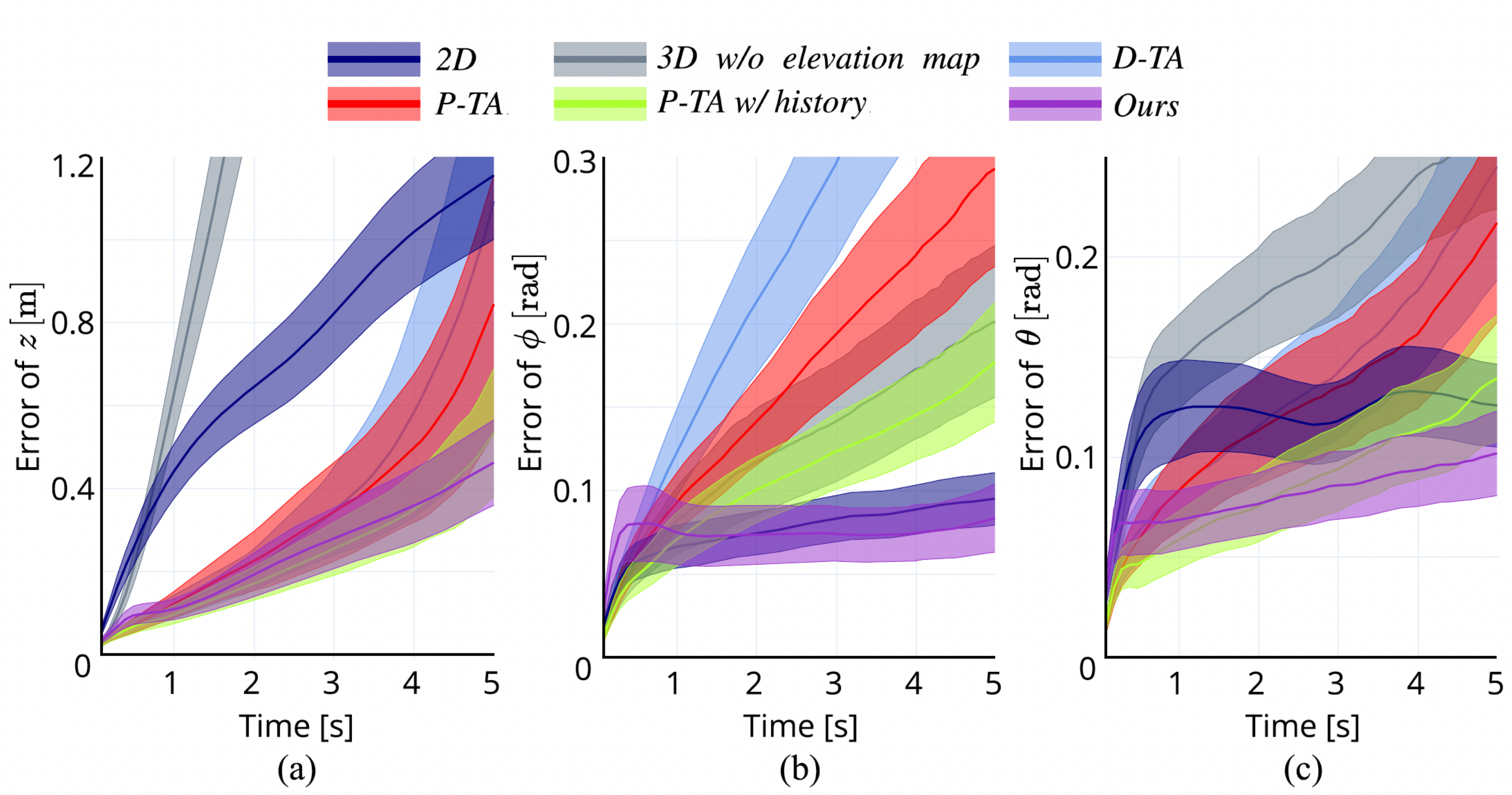}
\caption{Multi-step prediction errors for the values (a) $z$, (b) $\phi$, and (c) $\theta$. Each line represents the mean, and the shaded region represents $1/5$ of the standard deviation for visual clarity.}
\label{fig:multi_step_error}
\end{figure}

To address this issue, we simply generate trajectories by propagating the mean prediction~$\hat{\boldsymbol{\mu}}^{d}$. Prior work~\cite{chua2018deep} indicated that the performance of a controller using a mean propagation is comparable to that of a controller utilizing an MC method.

Based on the above strategy, we select the ensemble model that has the greatest aleatoric uncertainty, and then use the mean prediction of the chosen model. This enhances the robustness of the MPPI since it considers the worst-case scenario among the ensembles.

In summary, (\ref{Equation:dpnn}) is modified for use in MPPI as:
\begin{equation}  \hat{\mathbf{x}}^{d}_{t+1} = \mathbf{x}^{d}_{t} +     \hat{\boldsymbol{\mu}}^{d}_{l, t+1},\,
    l = \argmax\limits_{i \in \{1, 2, \dots, B\}} \left\lVert\hat{\boldsymbol{\sigma}}^{d}_{i, t+1}\right\rVert.
\label{Equation:imp_detail}
\end{equation}

\section{Experiments}
\subsection{Experimental Setup}
We conducted experiments using the high-fidelity vehicle simulation software, the IPG CarMaker. We collected driving data using a variety of human-selected maneuvers~\cite{kim2022smooth}. By leveraging a simulator, we could even include risky maneuvers such as sudden acceleration or changes in steering commands, which broadens the data distribution. We added Gaussian noise to $\delta$ and $v_{x}^{\text{des}}$ in order to enhance the accuracy of prediction when highly noisy control sequences arising from the MPPI sampling procedure are given to the model.

Two completely different elevation profiles were generated on the same track; one for training data collection and the other for model validation. We collected $200$ minutes of training data~$\mathcal{D}_{\text{train}}$ and $16$ minutes of validation data~$\mathcal{D}_{\text{val}}$.
\subsection{Training Procedure \label{subsection:extension}}
The dataset was preprocessed using min-max normalization and standardization of proprioceptive and exteroceptive information, respectively. We applied Gaussian noise with zero means to the data to improve model robustness~\cite{nagabandi2018neural}. Then we trained each model in the ensemble by minimizing the Gaussian negative log likelihood~(NLL) loss~$\mathcal{L}_{d}$:
\begin{equation}
\begin{aligned}
    \mathcal{L}_{d} &= \sum_{\mathbf{x}^{d} \in \mathcal{D}_{\text{train}}} \log\det \hat{\mathbf{\Sigma}} +  \hat{\bm{\eta}}^\top \hat{\mathbf{\Sigma}}^{-1}\hat{\bm{\eta}} \, ,
\end{aligned}
\end{equation}
where $\hat{\bm{\eta}} = \hat{\boldsymbol{\mu}}^{d}_{t+1}+\mathbf{x}^{d}_{t}-\mathbf{x}^{d}_{t+1}$, and $\hat{\mathbf{\Sigma}}$ is a diagonal matrix form of~$\hat{\boldsymbol{\sigma}}^{d}_{t+1}$.
\subsection{Analysis on The Kinodynamic Model}
\begin{figure*}[t]
\centering
\includegraphics[width=0.95\textwidth]{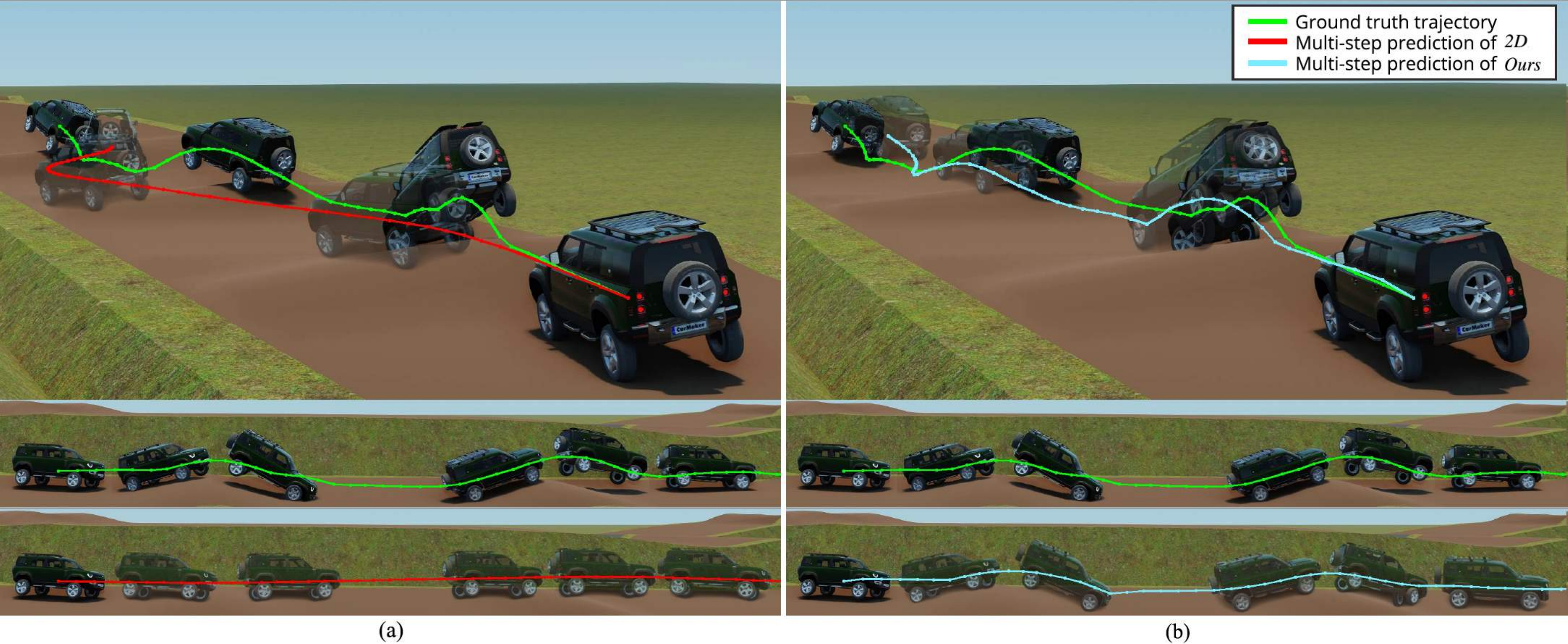}
\caption{Ground truth trajectory and predicted trajectories generated by multi-step prediction of (a) \textit{2D} and (b) \textit{Ours} for one example trial.}
\label{fig:example_traj}
\end{figure*}
We compare the prediction performances of kinodynamic models based on a number of design considerations explained in Section~\ref{section:dpnn}. Although we trained the model by minimizing the one-step loss, we chose~$T$-step validation error as the evaluation metric since we would like to know how well the MPPI controller can generate a trajectory by recursively applying the learned model~\cite{nagabandi2018neural}. Given the initial state~$\vx_{t}$ and the actual control sequence~$\vu_{t:t+T-1}$ from~$\mathcal{D}_{\text{val}}$, we compared the ground truth states~$\vx_{t+1:t+T}$ and the multi-step predictions of the model~$\hat{\mathbf{x}}_{t+1:t+T}$, which can be obtained by sequential one-step predictions:
\begin{equation}
\hat{\mathbf{x}}_{t+\tau} = 
\hat{\mathbf{x}}_{t+\tau-1} + F_{\bm{\theta}}\left(\hat{\mathbf{x}}_{t+\tau-1}, \mathbf{u}_{t+\tau-1}\right).
\end{equation}

To demonstrate the validity of our design considerations, we manually defined and trained the following ablation models using the same dataset: 1)~\textit{``2D"}: a 3-DoF baseline model; 2)~\textit{``3D w/o elevation map"}: a 6-DoF model without the knowledge of terrain; 3)~\textit{``D-TA"}: a Deterministic Terrain-Aware model; 4)~\textit{``P-TA"}: a Probabilistic Terrain-Aware model; 5)~\textit{``P-TA w/ history"}: a \textit{``P-TA"} provided with history information; 6)~\textit{``Ours"}: a \textit{``P-TA w/ history"} provided with orientation angles. In 1)-3), DPNN outputs the mean vector~$\hat{\boldsymbol{\mu}}^{d}$, and the mean squared error~(MSE) loss function is used for training. All models were ensemble models with a size of~$B = 5$. We list the design specifications in Table~\ref{tab:model_design}.
\begin{table}[ht]
\renewcommand\arraystretch{1.2}
\centering
\scriptsize
\caption{\small The design specifications of each model. We abbreviated the deterministic/probabilistic model as D/P, respectively. }
\resizebox{\linewidth}{!}{
    \begin{tabular}{ccccccc}
        \hline
        \multirow{2.5}{*}{\#}  & \multirow{2.5}{*}{Model} & \multicolumn{5}{c}{Input of DPNN} \\
        \cmidrule(l{4pt}r{4pt}){3-7} 
        & & DoF & $\mathbf{h}_t$ & D / P  & $H$ & $(\psi, \theta, \phi)$\\
        \hline
        1 & \textit{2D} & 3 & \xmark & D  & 1 & \xmark \\
        2 & \textit{3D w/o elevation map} & 6 & \xmark & D  & 1 & \xmark \\
        3 & \textit{D-TA} & 6 & \checkmark & D & 1 & \xmark \\
        4 & \textit{P-TA} & 6 & \checkmark & P & 1 & \xmark \\
        5 & \textit{P-TA w/ history} & 6 & \checkmark & P & 3 & \xmark \\
        6 & \textit{Ours}  & 6 &\checkmark & P & 3 & \checkmark \\
        \hline
    \end{tabular}
    }
    \label{tab:model_design}
\end{table}

Fig.~\ref{fig:multi_step_error} shows the multi-step prediction errors in the values~$(z, \theta, \phi)$ over the prediction horizon~$T$, which is highly dependent on terrain shape. \textit{2D} assumed constant values for these parameters, which resulted in discrepancies with ground truth values, and disregarded the influence of terrain geometry on the motion. \textit{3D w/o elevation map} exhibited more errors despite considering 6-DoF motion, due to the absence of terrain information, which led to divergence in a vertical direction. In contrast, \textit{D-TA} took advantage of both 6-DoF motion and the terrain geometry, making it capable of fully estimating 3D vehicle-terrain interactions. However, the result shows an eventual divergence on the open-loop prediction, due to the absence of uncertainty consideration.

\begin{table*}[t!]
\renewcommand\arraystretch{1.2}
\centering
\scriptsize
\caption{\small Experimental results on the race track. To assess the difficulty of the driving task, we conducted experiments using a baseline model at both $v_{\text{ref}} = 30$ km/h and $v_{\text{ref}} = 40$ km/h. We set $K$ and $T$ as 2000 and 20, respectively, which enables our algorithm to operate at 10Hz on the Nvidia RTX 3090 GPU with CUDA and PyTorch. We displayed the mean and standard deviation for lap time and $F_{z}^{\text{peak}}$.}
\resizebox{\textwidth}{!}{
\label{tab:results}
    \begin{tabular}{cccccccccccccc}
        \hline
        \multirow{2.5}{*}{\#}  & \multirow{2.5}{*}{Model} & \multirow{2.5}{*}{\shortstack{$v_{\text{ref}}$\\ (km/h)}} & \multicolumn{3}{c}{Off-Road Cost Functions} & \multirow{2.5}{*}{\shortstack{Lap Time\\ (s)}}  &
        \multirow{2.5}{*}{\shortstack{\# of\\ Failure}} &
        \multirow{2.5}{*}{\shortstack{$\left |v\right |_{\text{avg}}$\\ (km/h)}} &
        \multirow{2.5}{*}{\shortstack{$\left |v\right |_{\text{max}}$\\ (km/h)}} &
        \multirow{2.5}{*}{\shortstack{$\left |\phi\right |_{\text{max}}$\\ (deg)}} &
        \multirow{2.5}{*}{\shortstack{$\left |\theta\right |_{\text{max}}$\\ (deg)}} &
        \multirow{2.5}{*}{\shortstack{$F_{z}^{\text{peak}}$\\ (kN)}}\\
        \cmidrule(l{4pt}r{4pt}){4-6} 
        & & & $\text{Rollover}(\vx)$ & $\text{Force}(\vx)$ & $\text{UC}(\vx)$  & & & & & & & \\
        \hline
        1 & \textit{2D} & 30 & \xmark & \xmark & \xmark  & 127.62 $\pm$ 2.21 & 5 & 28.79 & 39.16  & 79.21 & 31.53 & 50.11 $\pm$ 20.17\\ 
        2 & \textit{2D} & 40 & \xmark & \xmark & \xmark  & 116.59 $\pm$ 4.76 & 19 & 32.56 & 42.22 & 79.41 & 36.90 & 53.23 $\pm$ 24.27\\
        3 & \textit{Ours} & 40 & \xmark & \xmark & \xmark & 117.45 $\pm$ 3.69 & 17 & 32.09 & 41.85 & 78.09 & 33.91 & 53.86 $\pm$ 21.96\\
        4 & \textit{Ours} & 40 & \checkmark & \xmark & \xmark & 120.86 $\pm$ 8.55 & 23 & 32.49 & 41.29 & 79.25 & 38.74 & 53.70 $\pm$ 26.44\\
        5 & \textit{Ours} & 40 & \checkmark & \checkmark & \xmark & 136.32 $\pm$ 2.07 & 3 & 28.04 & 41.37 & 78.71 & 76.42 & 44.50 $\pm$ 19.97\\
        \rowcolor{gray!50} 6 & \textit{Ours} & 40 & \checkmark &\checkmark & \checkmark & 134.69 $\pm$ 1.64 & 0 & 28.11 & 40.48 & 28.50 & 28.87 & 43.17 $\pm$ 16.22\\\hline
    \end{tabular}
    }
\label{Table:statistics_final}
\end{table*}

\begin{figure}[t]
\centering
\includegraphics[width=0.99\linewidth]{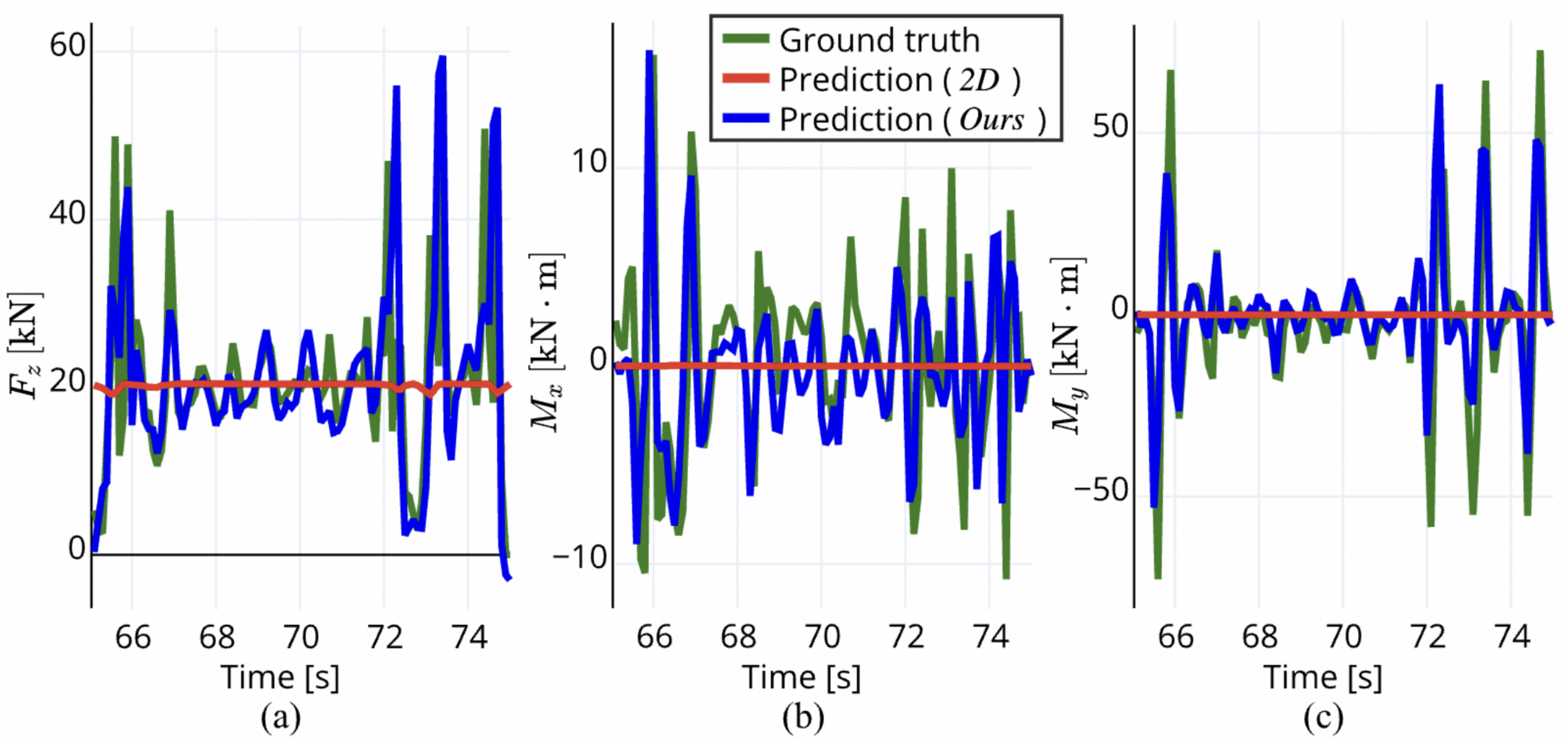}
\caption{Visualization of (a) vertical force~$F_{z}$, (b) roll moment~$M_{x}$, and (c) pitch moment~$M_{y}$ derived by using the predictions from \textit{2D} and \textit{Ours}.}
\label{fig:force_moment_prediction}
\end{figure}

The result of \textit{P-TA} indicates that incorporating stochasticity into a model reduces the accumulation of prediction errors. Additionally, integrating history information of states and control inputs provides further enhancements by capturing the time-varying characteristics of vehicle-terrain interaction, as shown by the result of \textit{P-TA w/ history}. Finally, our proposed model shows the most accurate multi-step predictions by using the orientation angles of the vehicle as input to the model. This is especially critical in terrains with severe curvatures or large altitude fluctuations, where changes in roll and pitch angles cannot be disregarded even if they are trivial in relatively flat terrain. Fig.~\ref{fig:example_traj} illustrates a multi-step prediction of \textit{2D} and \textit{Ours} at a specific time step.

We also visualize~$(F_{z},\,M_{x},\,M_{y})$ derived from~(\ref{Equation:force_and_moment}), using the predictions obtained from \textit{2D} and \textit{Ours} on~$\mathcal{D}_{val}$~(see Fig.~\ref{fig:force_moment_prediction}). The result indicates that the estimated forces and moments of our model closely align with the ground truth values obtained from force sensors in the simulator, without any training process involving ground truth force data. Notably, \textit{Ours} can reliably predict highly unsafe behavior in which the vehicle loses contact with the ground~($F_{z} = 0$) or lands again, causing a strong collision with the ground~($F_{z} \gg mg$). These findings validate the idea that integrating a force cost into the controller can help mitigate the risks resulting from severe impact with the ground.
\subsection{Autonomous Rally Driving Task}
The purpose of this section is to illustrate that both safe and aggressive driving over off-road terrain is achievable when the MPPI controller leverages information derived from our terrain-aware kinodynamic model. We evaluated the control performance of the MPPI using~\textit{Ours} with all the off-road costs, and manually defined other ablations by sequentially removing the following three off-road cost terms from~$c(\mathbf{x})$ for comparison: $\text{UC}(\mathbf{x})$,  $\text{Force}(\mathbf{x})$, and $\text{Rollover}(\mathbf{x})$. These ablations were performed to assess the contribution of each term to the control performance. The reference speed~$v_{\text{ref}}$ was set to $40$~km/h. Additionally, we evaluated the control performance of the MPPI using \textit{2D} at reference speeds of $30$~km/h and $40$~km/h. Note that it is infeasible to apply off-road costs when using \textit{2D}. This evaluation aimed to establish a lower bound baseline and assess the difficulty of high-speed driving on off-road terrain.

The objective of each controller was to complete 10 counterclockwise laps around the track. In these experiments, we employed the same race track that was used for validation data collection. If the vehicle deviated from the track or significantly lost its stability, we repositioned it at the centerline of the track and continued driving for the remaining laps. To evaluate the controllers, we measured the maximum roll and pitch angles of the vehicle and the range of impulsive vertical force caused by shock impacts with the ground.

 Table~\ref{Table:statistics_final} presents the experimental results. Results related to \textit{2D} demonstrates that ignoring the terrain geometry can lead to catastrophic failures, such as overturning, and frequent loss of ground contact, hence increasing both the frequency and magnitude of collisions with the ground. Similar tendencies are observed when using \textit{Ours} without any off-road cost terms, or when using only the rollover cost in~$c(\mathbf{x})$. This is because the orientation angles of the vehicle are insufficient to fully identify unsafe conditions such as strong ground impacts. We observe that the addition of the force cost considerably reduces the intensity of impulsive impact on the vehicle and lowers the frequency of failures. Nonetheless, failure still occurred because the estimated state cost based on model prediction does not consider the uncertainty. Finally, the inclusion of the uncertainty cost term guided the vehicle to a region with less uncertainty, resulting in the completion of 10 laps without any failure and a reduced magnitude of impulsive force.
 
\begin{figure}[t]
\centering
\includegraphics[width=0.99\linewidth]{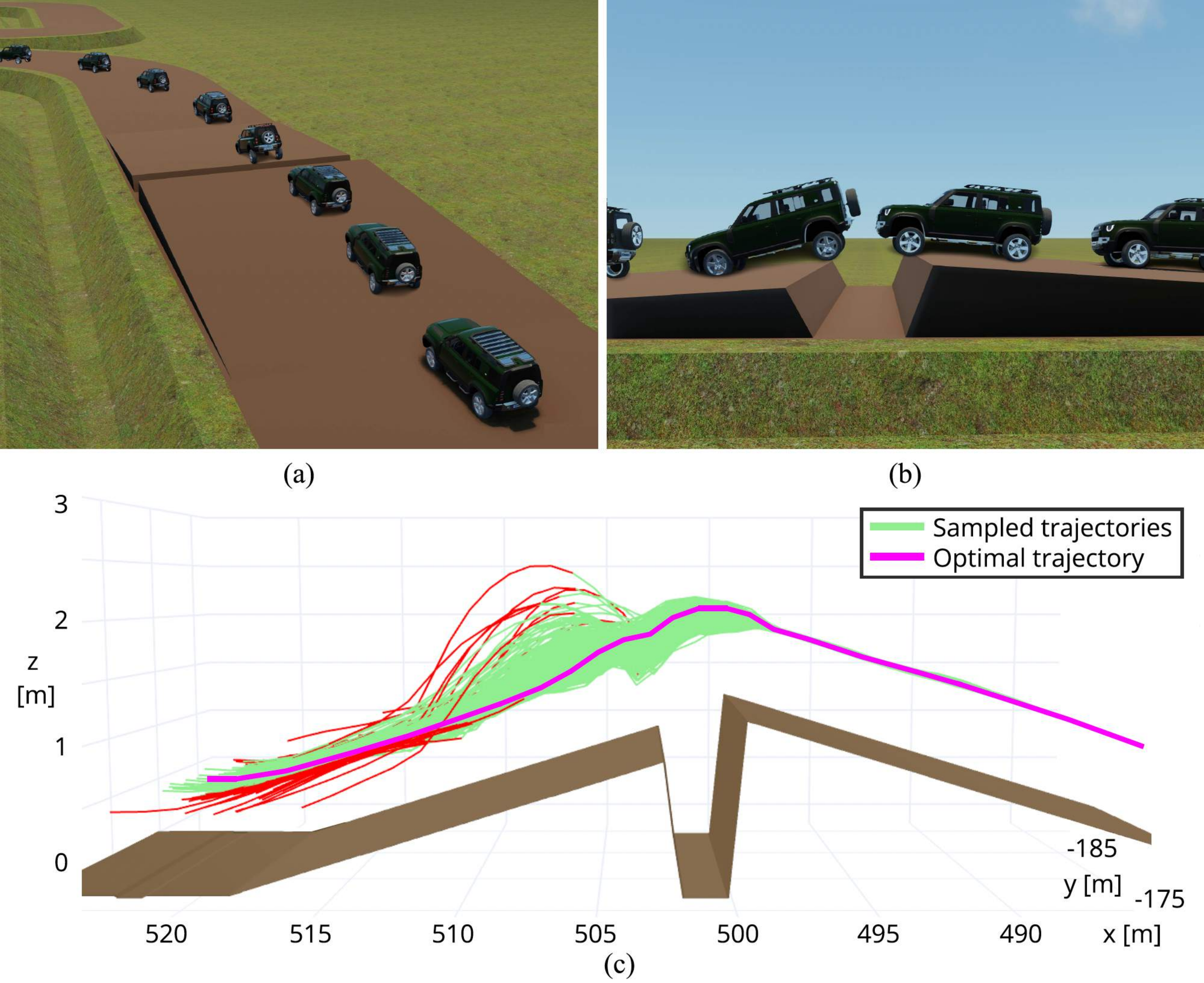}
\caption{(a) The back view, and (b) the side view of actual vehicle trajectory while crossing a cracked region. (c) The sampled trajectories of the MPPI right before the vehicle crosses the crack and the computed optimal trajectory in that time step. Intriguingly, it is estimated that trajectories with low speeds at which cannot be crossed are subjected to a significant impact on the vehicle's rear wheels by the terrain, resulting in a loss of stability (colored in red).}
\label{fig:edge_case}
\end{figure}

Lastly, we illustrate a specific scenario where our model facilitates the controller to execute more aggressive maneuvers in order to traverse a challenging terrain. Fig.~\ref{fig:edge_case} exhibits the control outcomes of the MPPI using \textit{Ours} when crossing a cracked area on the terrain that cannot be traversed at low speeds. Our terrain-aware model is able to predict the impact by the cracked road, including a complete loss of stability at relatively low speeds. Hence, the controller slightly accelerates the vehicle to successfully cross the cracked region. In contrast, \textit{2D} baseline stops the vehicle in front of the crack, judging that it is a region that cannot be traversed. This result demonstrates that the use of our terrain-aware model in the controller improves the vehicle's maneuverability by considering the distinctive features of the terrain, instead of adopting a conservative approach by simply lowering the speed and cautiously driving over challenging terrain.

\section{Conclusion}
We presented a 6-DoF terrain-aware kinodynamic model for high-speed vehicle control on off-road terrain. Our model shows reliable predictions of vertical, roll, and pitch motions closely related to stability and physical feasibility, by using proprioceptive and exteroceptive information as inputs along with a few design considerations. Additionally, our model can predict contact interactions in terms of forces and moments, allowing for assessments of ground impact severity without any training using ground truth force data. Our MPC can utilize these predictions to execute safe and aggressive control inputs, leading to improved vehicle maneuverability. Our experiments demonstrate that our model-controller pair outperforms the baseline in terms of safety and robustness. Our future research will extend this work by applying this model to real-world vehicles via sim-to-real transfer.

\bibliographystyle{IEEEtran}
\typeout{}
\bibliography{references.bib}

\end{document}